\documentclass{article}
\usepackage{spconf,amsmath,graphicx,hyperref,amsfonts,amsthm,enumitem,mathtools,cite,caption,subcaption,float, lineno}

\def\F{{\mathbf F}}
\def\f{{\mathbf f}}
\def\W{{\mathbf W}}

\def\z{{\mathbf z}}

\def\R{\mathbb{R}}
\def\I{{\mathbf I}}
\def\b1{{\mathbf 1}}

\newtheorem{proposition}{Proposition}
\newtheorem{assumption}{Assumption}

\title{Learning Beyond the Gaussian Data: Learning Dynamics of Neural Networks on an Expressive and Cumulant-Controllable Data Model}
\name{Onat Üre, Samet Demir, Zafer Doğan\thanks{We acknowledge that this work is supported partially by TÜBİTAK under project 124E063 in ARDEB 1001 program. O.Ü. is supported by an MS Scholarship (BİDEB 2210) from TÜBİTAK and an AI Fellowship provided by KUIS AI Research Center. S.D. is supported by an AI Fellowship provided by KUIS AI Research Center and a PhD Scholarship (BİDEB 2211) from TÜBİTAK. The corresponding author is Zafer Doğan (zdogan@ku.edu.tr).}}
\address{College of Engineering and KUIS AI Center, Koç University, Istanbul, Turkey\\
\{oure25, sdemir20, zdogan\}@ku.edu.tr}
\begin{document}
\maketitle

\begin{abstract}
We study the effect of high-order statistics of data on the learning dynamics of neural networks (NNs) by using a moment-controllable non-Gaussian data model. Considering the expressivity of two-layer neural networks, we first construct the data model as a generative two-layer NN where the activation function is expanded by using Hermite polynomials. This allows us to achieve interpretable control over high-order cumulants such as skewness and kurtosis through the Hermite coefficients while keeping the data model realistic. Using samples generated from the data model, we perform controlled online learning experiments with a two-layer NN. Our results reveal a moment-wise progression in training: networks first capture low-order statistics such as mean and covariance, and progressively learn high-order cumulants. Finally, we pretrain the generative model on the Fashion-MNIST dataset and leverage the generated samples for further experiments. The results of these additional experiments confirm our conclusions and show the utility of the data model in a real-world scenario. Overall, our proposed approach bridges simplified data assumptions and practical data complexity, which offers a principled framework for investigating distributional effects in machine learning and signal processing.
\end{abstract}

\section{Introduction}
\label{sec:intro}

Modern datasets—including images and speech to sensor measurements—exhibit complex statistical dependencies. In practice, data are rarely Gaussian: multimodality, skewness, heavy tails, and high kurtosis are common and strongly shape the learning behavior. Yet, most theoretical analyses of generalization rely on Gaussian assumptions~\cite{pmlr-v235-cui24d,pmlr-v235-moniri24a,10.5555/3600270.3603019, demir2025asymptotic}. These assumptions offer mathematical convenience but fundamentally miss the high-order structures that drive performance in real-world applications.

High-order statistics described by cumulants (an alternative to the moments while being a measure of non-Gaussianity) \cite{novak2014three, fisher1932derivation} are highly relevant to learning dynamics. Specifically, understanding how high-order statistics shape generalization has therefore become a priority~\cite{doi:10.1073/pnas.2201854119,10.5555/3618408.3619607,PhysRevResearch.5.033177,belrose,NEURIPS2024_8f8af4ee,bardone,  riccifeature25}. Recent evidence shows that NNs trained with stochastic gradient descent do not harness all statistical features simultaneously; instead, they first align with low-order moments (mean and covariance) before leveraging high-order cumulants~\cite{doi:10.1073/pnas.2201854119, 10.5555/3618408.3619607, PhysRevResearch.5.033177, belrose}. Several synthetic models attempt to capture this staged behavior. For instance, spiked-cumulant models~\cite{NEURIPS2024_8f8af4ee} and mixed-cumulant models~\cite{bardone} inject structured non-Gaussianity into Gaussian data to enable controlled experiments. However, these approaches remain fundamentally limited because they hard-code specific non-Gaussian structures rather than offering broad, tunable control.

This paper addresses this gap by introducing a general and analytically tractable generative framework for modeling non-Gaussian data. Our approach enables fine-grained control over high-order cumulants (including skewness and kurtosis) and reveals their impact on network generalization. Concretely, we construct a two-layer generative neural network in which the activation function is expanded via Hermite polynomials~\cite{O’Donnell_2014}. By tuning the Hermite coefficients, we systematically modulate the degree and type of non-Gaussianity. This design allows, for the first time, a clean disentanglement of individual cumulants and their contributions to learning dynamics, closing the gap between simplistic Gaussian-based analyses and the rich distributions encountered in practice.

Our main contributions are as follows:
(i) We introduce a framework that offers an independent control of high-order cumulants (e.g., skewness, kurtosis) while remaining expressive enough to represent broad families of non-Gaussian distributions. This provide a powerful new tool for systematically studying non-Gaussian effects in machine learning and signal processing.
(ii) Through controlled experiments on synthetic and real-world datasets, we show that NNs trained on our model exhibit moment-wise learning dynamics, progressively incorporating high-order statistics during training.

\section{Proposed Framework}
\label{sec:problem}

We propose a framework for systematically studying the impact of high-order cumulants on the learning dynamics of a two-layer neural network.  
At its heart is a generative model in which non-Gaussianity is introduced through a tunable \textit{nonlinear mapping}.  
Rather than injecting non-Gaussianity ad hoc, we treat this map as a small neural module whose Hermite coefficients act as “dials” to adjust skewness, kurtosis, and other cumulants.  
This yields a simple and analytically tractable mechanism to turn high-order effects on or off and observe their impact on downstream learning.

In practice, generative models --- such as Generative Adversarial Networks (GANs) \cite{goodfellow2014generative}, Normalizing Flows \cite{kobyzev2020normalizing}, and Diffusion models \cite{yang2023diffusion} --- are generally constructed by using a latent Gaussian vector $\z$ together with a nonlinear mapping $\Theta(\mathbf{z})$ that converts Gaussian vector to a non-Gaussian vector. Motivated by this, we follow the same approach. Formally, let $\mathbf{z}\sim\mathcal{N}(\boldsymbol{\mu},\boldsymbol{\Sigma})\in\mathbb{R}^p$ be a latent Gaussian vector. In this work, we focus on the following \textit{nonlinear mapping}:
\begin{equation}
  \Theta(\mathbf{z};\mathbf{F}) := \sigma(\mathbf{F}\mathbf{z}) \in \mathbb{R}^k,
  \label{theta_def}
\end{equation}
where \(\mathbf{F}\in\mathbb{R}^{k\times p}\) denotes the parameters of the mapping  and \(\sigma:\mathbb{R}\to\mathbb{R}\) a nonlinear activation. Then, a generative data model takes the compact form
\begin{equation}
  \hat{\mathbf{x}} := \mathbf{W}\,\Theta(\mathbf{z};\mathbf{F}) \in \R^d,
  \label{data_model}
\end{equation}
where \(\mathbf{W}\in\mathbb{R}^{d\times k}\) represents the final linear layer. Here, note that all non-Gaussianity arises solely from \(\Theta\).

To parameterize and control the high-order cumulants of \(\Theta(\cdot)\), we expand each component of \(\Theta\) in Hermite polynomials~\cite{O’Donnell_2014}, which form a complete orthogonal basis under the Gaussian measure, supposing the following assumption.

\begin{assumption}
For every row \(\f_i^\top\) of \(\mathbf{F}\), the component \(\sigma(\f_i^\top\mathbf{z})\) satisfies
\(\mathbb{E}_{\mathbf{z}}\bigl[\sigma(\f_i^\top\mathbf{z})^2\bigr]<\infty.\)
\end{assumption}

Under this assumption, each component of \(\Theta\) admits the Hermite expansion
\begin{equation}
  \Theta(\mathbf{z};\mathbf{F})
  = \lim_{\ell\to\infty}\sum_{i=0}^{\ell} c_i\,\mathrm{He}_i(\mathbf{F}\mathbf{z}),
  \label{hermite_exp_theta}
\end{equation}
where \(\mathrm{He}_i(\cdot)\) denotes the \(i\)-th probabilist's Hermite polynomial and \(c_i\) its expansion coefficient. 
In the Hermite expansion, each coefficient \(c_i\) governs the strength of the associated  cumulant in the polynomial representation, thus enables direct and interpretable control over the statistical structure of the generated data.

\begin{proposition}
[Expressivity]
\label{prop:prop1}
Suppose \eqref{hermite_exp_theta} holds and \(p>d\). Then, the class of data models defined by \eqref{data_model} is dense in the space of \(d\)-dimensional probability measures with finite moments under standard weak topologies (e.g., Wasserstein metrics).
\begin{proof}[Proof (Concept)]
By the universal approximation theorem, any sufficiently regular function can be approximated by a two-layer neural network~\cite{ctx1845585390009246}.  
Hence, the class defined by \eqref{data_model} is dense in the space of \(d\)-dimensional probability measures with finite moments (e.g., under the Wasserstein metric). 
Since Hermite polynomials form a complete orthogonal basis of square-integrable functions under the Gaussian measure~\cite{O’Donnell_2014}, as \(\ell\!\to\!\infty\) any activation \(\sigma\), and thus any featurization map \(\Theta\) applied to Gaussian inputs, admits an expansion of the form \eqref{hermite_exp_theta} without loss of expressivity. Consequently, by tuning the Hermite coefficients and network parameters, the induced model can approximate any \(d\)-dimensional target with finite moments in the chosen weak/Wasserstein sense.
\end{proof}

\end{proposition}

Thus, \eqref{data_model} yields a cumulant-controllable yet highly expressive data model for probing non-Gaussian effects on learning.  
Moreover, one does not need infinitely many coefficients \(\{c_i\}_{i=0}^\infty\) to achieve precise cumulant control:

\begin{proposition}[Controllability of cumulants for finite \(\ell\)]
\label{prop:prop2}
For finite \(\ell\), given \(\ell+1\) target cumulants, one can always find \(\ell+1\) Hermite coefficients achieving them.
\begin{proof}
Let \(\sigma_\ell(z):=\sum_{i=0}^{\ell} c_i\mathrm{He}_i(z)\) denote the truncated expansion with \(\ell+1\) coefficients.  
Because \(\mathbf{W}\) in \eqref{data_model} acts as an affine transform, it suffices to analyze the cumulants of \(\sigma_\ell(\mathbf{F}\mathbf{z})\).  
Without loss of generality, consider the scalar case \(\sigma_\ell(\f^\top\mathbf{z})\) with \(\f\in\mathbb{R}^p\).  
Let \(\kappa_i\) denote its \(i\)-th cumulant.  
Then there exist polynomial maps \(g_i:\mathbb{R}^{\ell+1}\to\mathbb{R}\) such that \(\kappa_i=g_i(c_0,\dots,c_\ell)\).  
Given \(\{\kappa_i\}_{i=0}^\ell\), solving the resulting system of \(\ell+1\) polynomial equations yields the required coefficients \(\{c_i\}_{i=0}^\ell\).  
Therefore, cumulants up to order \(\ell\) can be set exactly via the first \(\ell+1\) Hermite coefficients.
\end{proof}
\end{proposition}

Since controlling \((\ell+1)\) cumulants only requires the \(\ell\)-th order Hermite expansion—i.e., \((\ell+1)\) coefficients \(\{c_i\}_{i=0}^{\ell}\)—we restrict attention to the following finite-order generative model for the remainder of this work:
\begin{equation}
    \mathbf{x} = \mathbf{W}\Theta_\ell(\mathbf{z};\mathbf{F}),
  \label{final_data_model}
\end{equation}
where $\Theta_\ell$ is $\ell$-th order polynomial mappping defined as $\Theta_\ell(\z, \F) := \sigma_\ell(\F \z)$ where $\sigma_\ell(z):=\sum_{i=0}^{\ell} c_i\mathrm{He}_i(z)$ with \(c_i\in\mathbb{R}\) coefficients used to control the cumulants.

By selecting nonzero coefficients \(c_i\) for \(i\geq2\), we explicitly introduce high-order cumulants into the data distribution.  
Thanks to the orthogonality of Hermite polynomials, each cumulant can be estimated independently without interference from other polynomial terms.  
Although each coefficient contributes to several cumulants, the mapping remains direct and interpretable, giving principled and fine-grained control over how the distribution departs from Gaussianity.

\begin{figure*}[t]
    \centering    
    \begin{subfigure}{0.24  \textwidth}
    \centering
    \includegraphics[width=\linewidth]{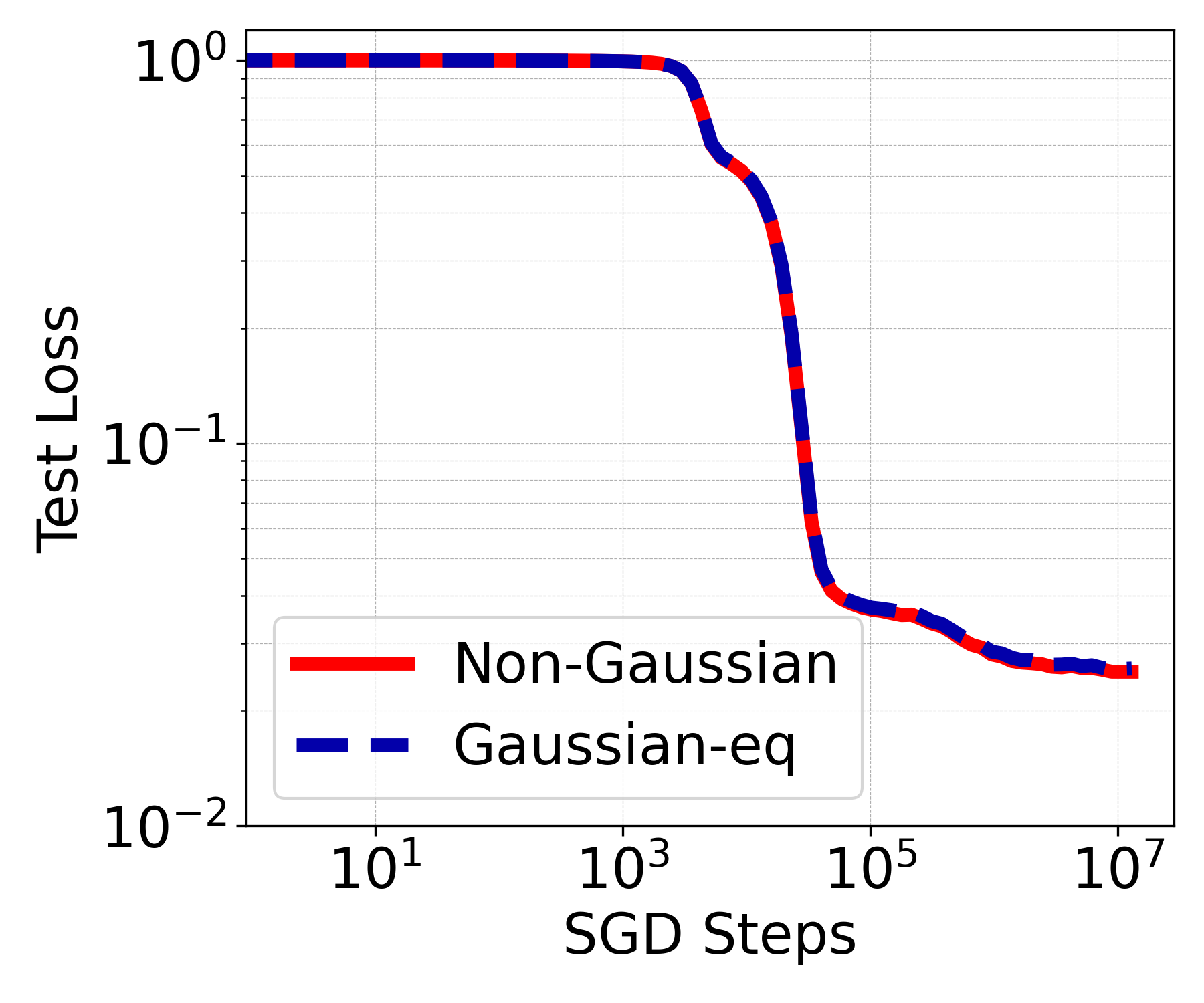}
    \caption{\centering Gaussian data without any high order cumulants\\($c_2=c_3=0$)}
    \label{fig:subfig1}
    \end{subfigure}
    \begin{subfigure}{0.24\textwidth}
        \centering
        \includegraphics[width=\linewidth]{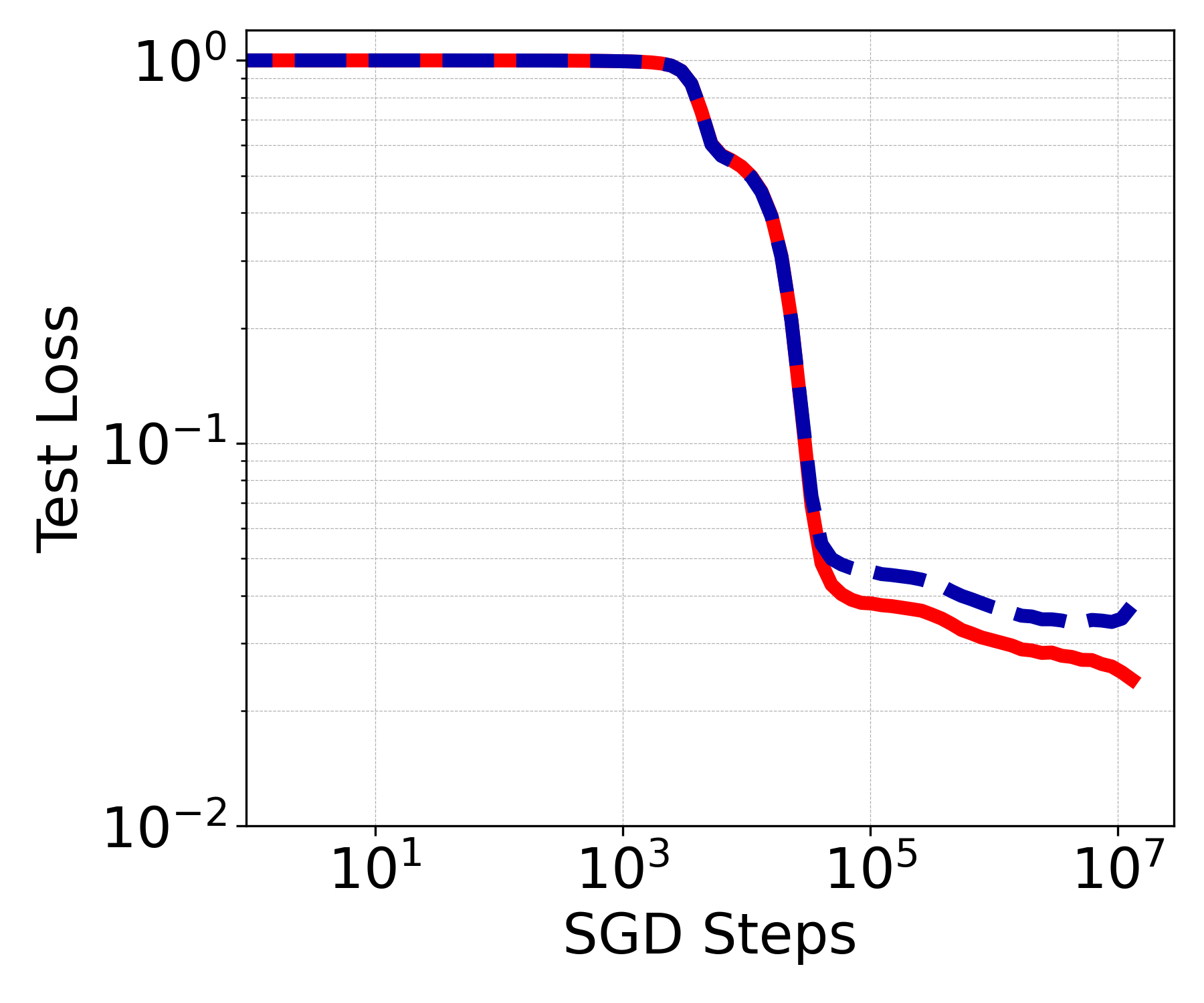}
        \caption{\centering Non-Gaussian data with high order cumulants from 2nd coeff.\\($c_2 = 0.2>0, c_3=0$)}
        \label{fig:subfig2}
    \end{subfigure}
    \begin{subfigure}{0.24\textwidth}
        \centering
        \includegraphics[width=\linewidth]{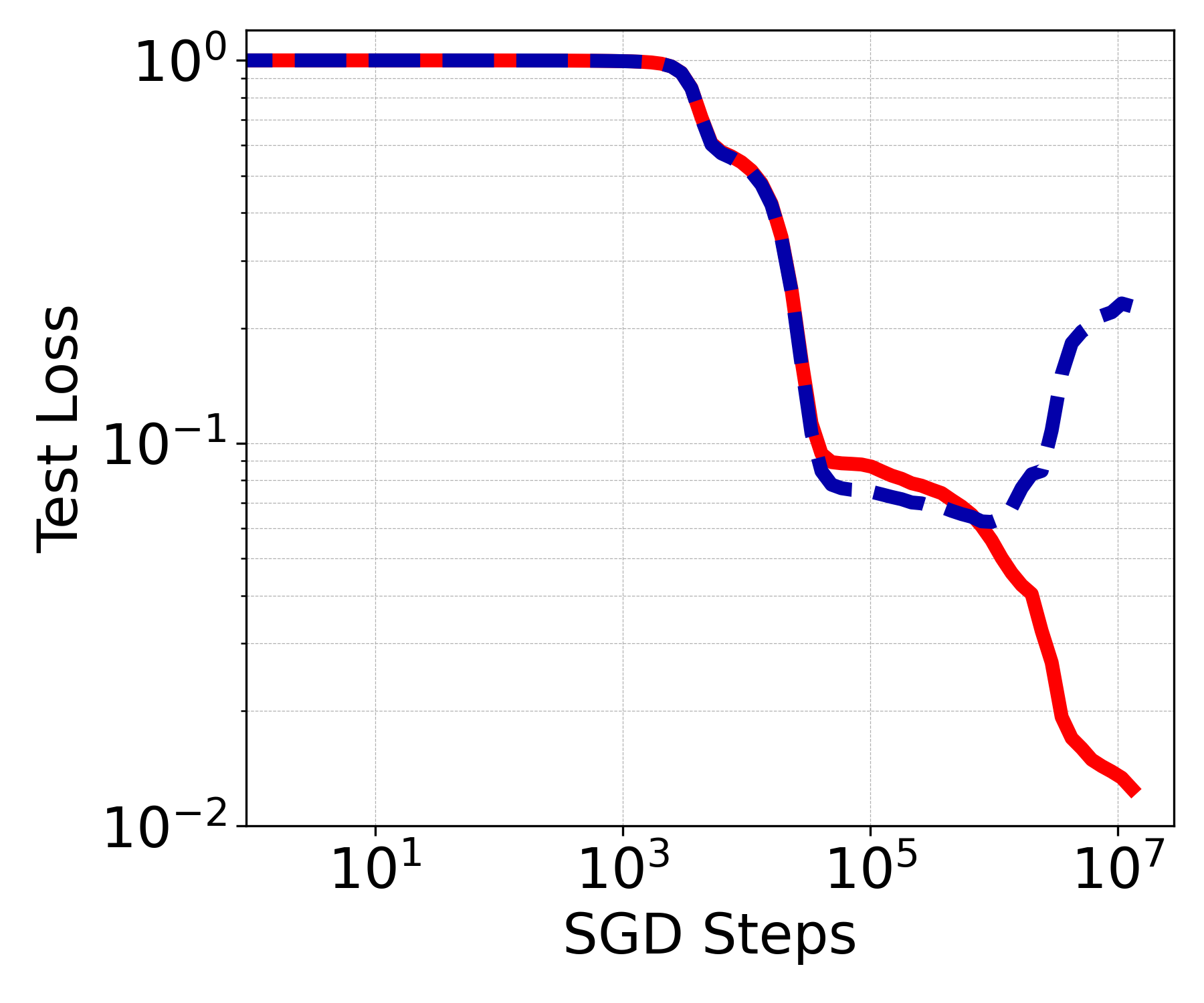}
        \caption{\centering Non-Gaussian data with high order cumulants from 3rd coeff.\\($c_2=0, c_3 = 0.2 >0$)}
        \label{fig:subfig3}
    \end{subfigure}
    \begin{subfigure}{0.24\textwidth}
        \centering
        \includegraphics[width=\linewidth]{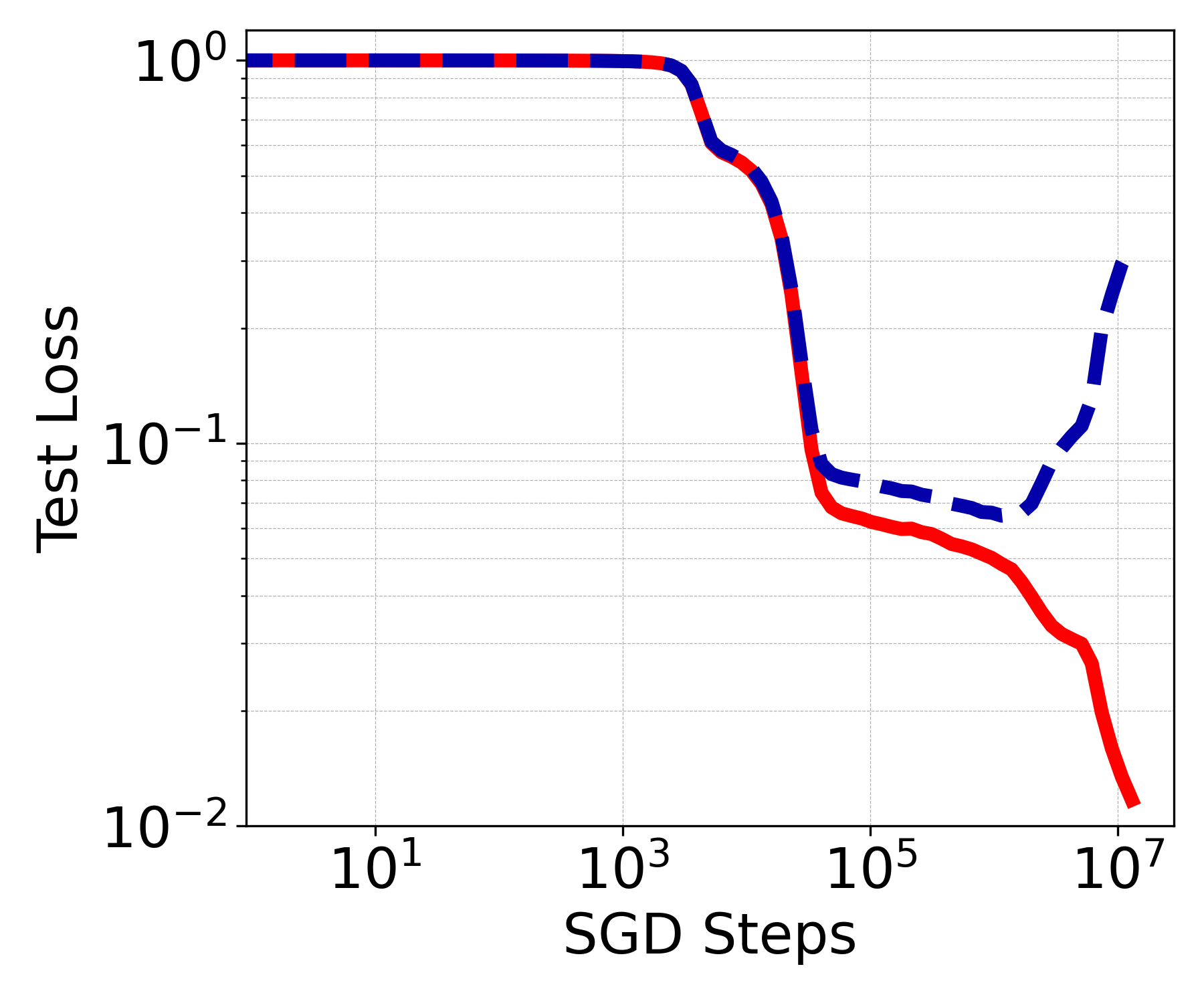}
        \caption{\centering Non-Gaussian data with high order cumulants from both coeff.\\($c_2 = 0.2>0, c_3 = 0.2>0$)}
        \label{fig:subfig4}
    \end{subfigure}
    \caption{
     \textit{Gaussian vs. non-Gaussian binary classification (synthetic data):} Test losses on the non-Gaussian dataset and its Gaussian-equivalent counterpart with matched mean and covariance (Sec~\ref{sec:exp-setting}) for a two-layer neural network trained to discriminate between standard Gaussian and non-Gaussian data \eqref{final_data_model}. For the non-Gaussian data model \eqref{final_data_model}, we set parameters $\mathbf{W} = \mathbf{F} = \mathbf{I}$, $\boldsymbol{\mu} = \mathbf{0}$, $\boldsymbol{\Sigma} = \I_p$, while the degree of Hermite expansion is $\ell = 3$, the latent dimension is $p = 128$ and the Hermite coefficients are chosen as $c_0 = 0.4$, and $c_1 = 0.5$. The considered two-layer neural network has $512$ hidden units and ReLU as activation function. The network is optimized using online SGD with mean-squared error loss, and the learning rate of $10^{-1}$. Results are averaged over 5 independent experiments.
    }
    \label{fig:c2c3}
\end{figure*}
\section{EXPERIMENTAL RESULTS}
In this section, we present experimental results evaluating how our cumulant-controllable non-Gaussian data model affects generalization. To directly assess the influence of high-order cumulants, we conduct a series of experiments on both synthetic and real datasets.
\subsection{Experimental setting}

\label{sec:exp-setting}
We illustrate the role of high-order statistics in the learning
process by introducing cumulant-controllable data model, which allows us to isolate the effect of non-Gaussian structure on the
generalization behavior of a network. In particular, the Hermite expansion coefficients provide direct and interpretable control over the distributional structure of the generated data, independent of any specific downstream task. 
To demonstrate this capability, in our experiments, we consider a binary classification problem in which the two classes differ in their distributions, described as folows:

\textbf{Non-Gaussian dataset --} 
\label{sec:non-gauss-setting}
We construct a balanced dataset of $2N$ samples, with $N$ examples per class. Class distributions are as follows: The first class (label $-1$) consists of samples drawn from a standard Gaussian distribution, while the second class (label $+1$) is composed of non-Gaussian samples generated via our data model \eqref{final_data_model}. The key distinction lies in the high-order statistics. While the distribution of the first class possesses only first- and second-order structure (with all high-order cumulants vanishing due to Gaussianity), the distribution of the second class exhibits rich high-order dependencies introduced by the nonlinear data model \eqref{final_data_model}. 
\textbf{Gaussian-equivalent evaluation dataset --}
\label{sec:Gauss-eq-setting}
In parallel, to disentangle the role of high-order statistics in generalization, we construct an auxiliary \emph{Gaussian-equivalent dataset} where the first class stays the same, while the second class is constructed by sampling from a Gaussian distribution with first and second moments matched with those of our data model \eqref{final_data_model}. This allows us to isolate the specific contribution of high-order cumulants to both learning dynamics and generalization by comparing the test performance on the non-Gaussian dataset versus the Gaussian equivalent dataset.
\textbf{Training procedure --} In the experiments, we train a neural network model on the non-Gaussian dataset and evaluate its generalization performance on both the non-Gaussian and the Gaussian-equivalent datasets, illustrating the impact of the high-order statistics on the learning dynamics.  

\begin{figure*}[t]
    \centering
    \begin{subfigure}{0.24\textwidth}
        \centering
        \includegraphics[width=\linewidth]{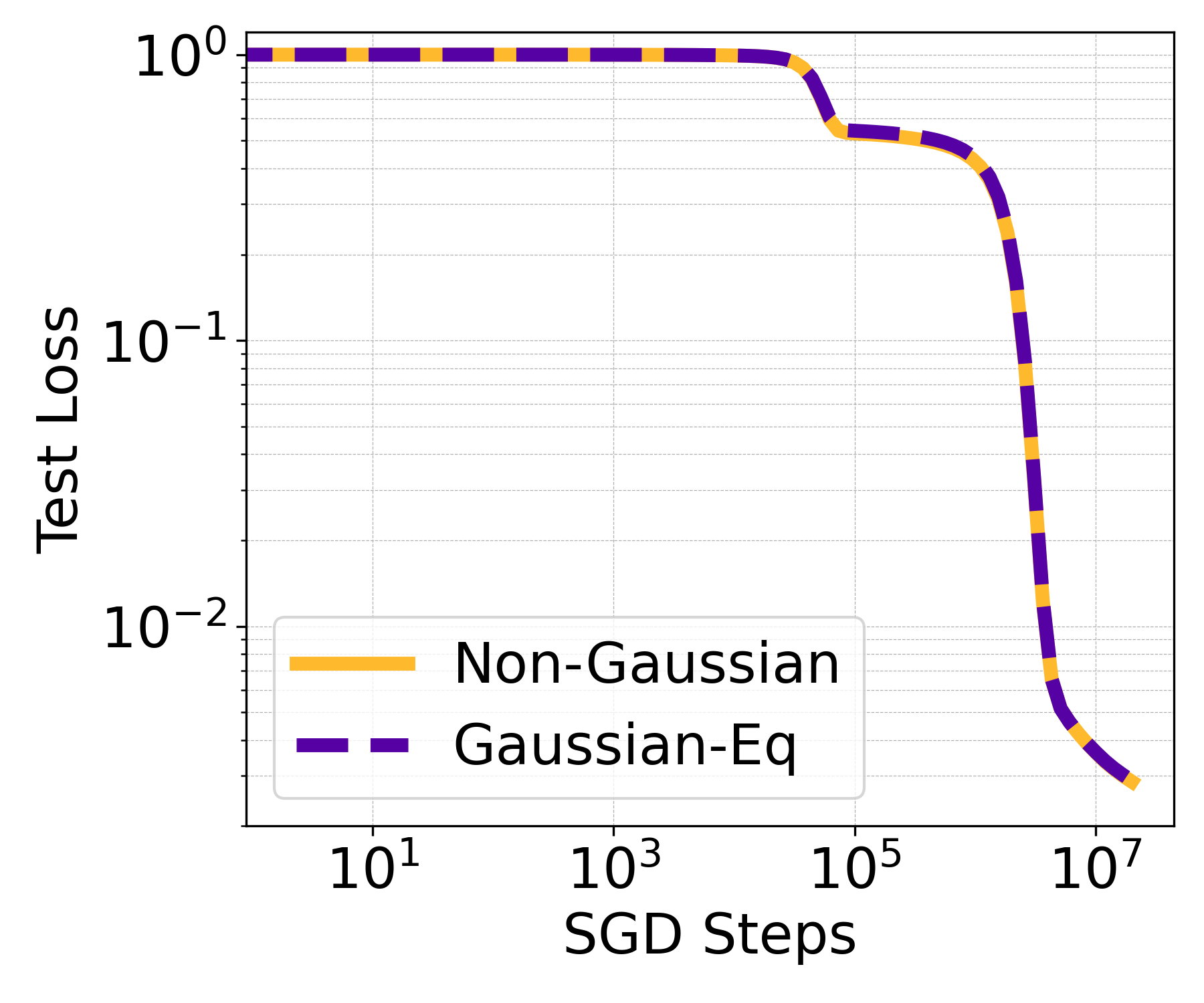}
        \caption{$c_3=c_5=0$}
        \label{fig:fashion_subfig1}
    \end{subfigure}%
    \hspace{0.001em}
    \begin{subfigure}{0.24\textwidth}
        \centering
        \includegraphics[width=\linewidth]{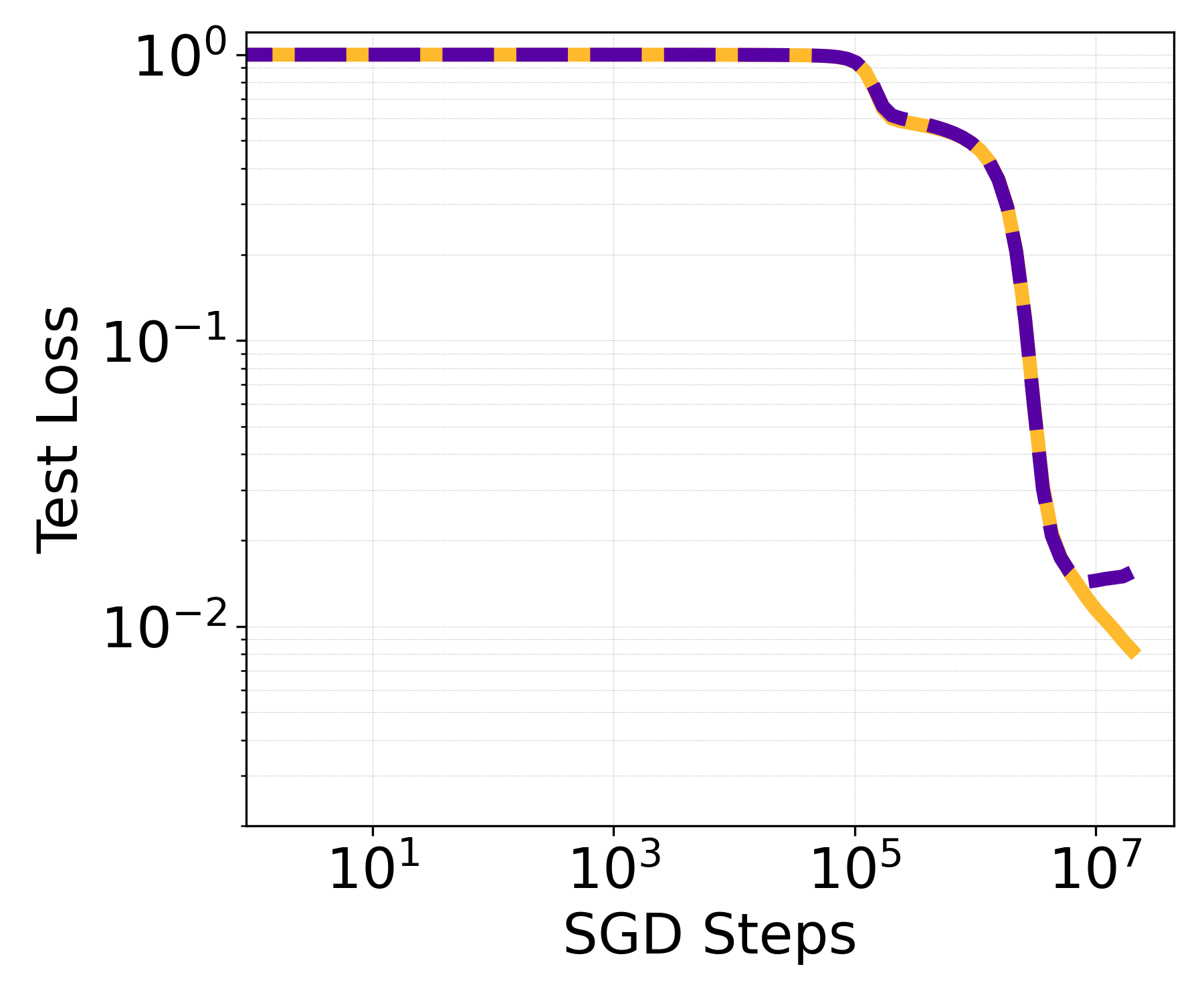}
        \caption{$c_3 \neq 0, c_5=0$}
        \label{fig:fashion_subfig2}
    \end{subfigure}%
    \hspace{0.001em}
    \begin{subfigure}{0.24\textwidth}
        \centering
        \includegraphics[width=\linewidth]{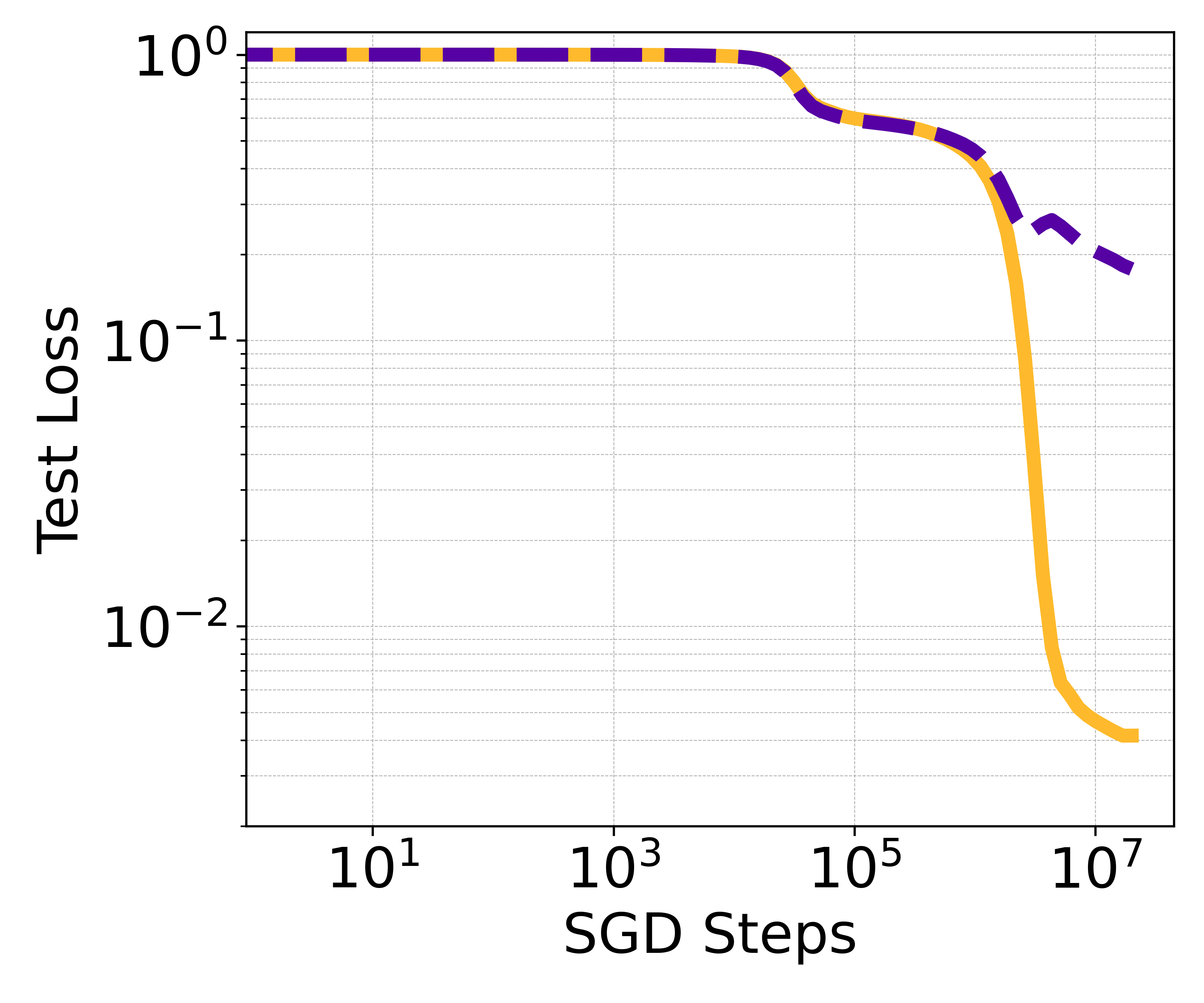}
        \caption{$c_3=0, c_5 \neq 0$}
        \label{fig:fashion_subfig3}
    \end{subfigure}%
    \hspace{0.001em}
    \begin{subfigure}{0.24\textwidth}
        \centering
        \includegraphics[width=\linewidth]{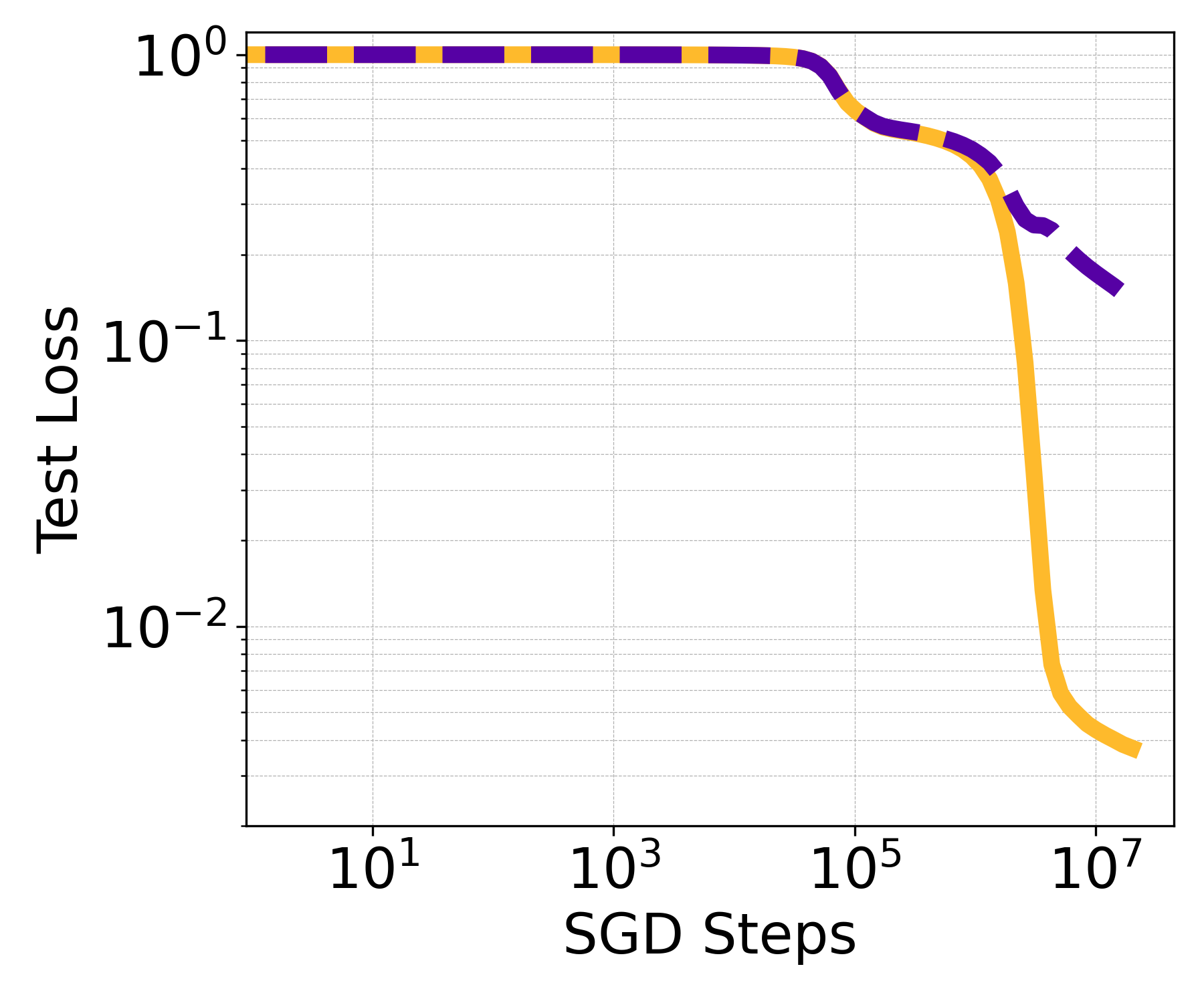}
        \caption{$c_3\neq0, c_5\neq0$}
        \label{fig:fashion_subfig4}
    \end{subfigure}
    \caption{
     \textit{Gaussian vs. non-Gaussian binary classification (\textsc{Fashion-MNIST} data):} Test losses on the non-Gaussian dataset and its Gaussian-equivalent counterpart  with matched mean and covariance (Sec~\ref{sec:exp-setting} and ~\ref{sec:mnist}) for a two-layer neural network trained to discriminate between standard Gaussian and non-Gaussian \textsc{Fashion-MNIST} samples generated by our pretrained data model \eqref{final_data_model}. For this experiment, we focus only on the samples of the T-shirt/top class (label 0). To train the parameters of the data model, we consider a GAN model with a generator of the form $\W \tanh(\mathbf{Fz} + \mathbf{b})$, where only $\mathbf{F}$ and $\mathbf{b}$ are trained to produce realistic T-shirt samples while $\W$ is set to identity and a discriminator is a two-layer ReLU network with a final sigmoid output. Both generator and discriminator are trained adversarially for 100 epochs using Adam (learning rate $10^{-4}$, $\beta_1=0.9$, $\beta_2=0.999$), batch size 48, and binary cross-entropy loss. After the training of the GAN \cite{goodfellow2014generative}, we apply Hermite expansion (truncated at degree $\ell =5$) to $\tanh$ and use the resulting Hermite coefficients together with the parameters $\W, \F, \mathbf{b}$ to reach our cumulant-controllable data model \eqref{final_data_model}. For the binary classification, the two-layer neural network is considered with $512$ hidden units, and the latent dimension is $p =100$. The model is optimized via online stochastic gradient descent (SGD) with a mean squared error loss and a fixed learning rate of $3 \times 10^{-4}$.
    }    
    \label{fig:fashion}
\end{figure*}
    
 \subsection{Experiments with synthetic data}
To focus on the effects of non-Gaussianity (arising solely from the nonlinearity) on the learning dynamics, we begin with our synthetic data experiments where the parameter matrices of the data model \eqref{final_data_model} are set to identity. Specifically, the input data are constructed using a third-order truncation ($\ell = 3$) of the Hermite expansion in \eqref{final_data_model}. By fixing $c_0$ and $c_1$, we maintain consistent low-order structure and use $c_2$ and $c_3$ to modulate high-order polynomials; thereby, high-order cumulants such as skewness, and kurtosis.

The behavior of test losses under varying high-order Hermite coefficients is illustrated in Fig.~\ref{fig:c2c3}. This result confirms that the non-Gaussianity can be controlled with finitely many coefficients $c_i$ as anticipated by Proposition~\ref{prop:prop2}. Specifically, when the high-order coefficients vanish ($c_{2}=c_{3}=0$ as in Fig.~\ref{fig:c2c3}a), the data model \eqref{final_data_model} reduces to pure Gaussian. Then, introducing a nonzero $c_{2}$ (as in Fig.~\ref{fig:c2c3}b) modifies the distribution by adding skewness and altering kurtosis, thereby enriching the data with additional statistical structure beyond the first two moments. In contrast, activating $c_{3}$ (as in Fig.~\ref{fig:c2c3}c-d) contributes an antisymmetric component, since $\mathrm{He}_3$ is an odd function. This injects sign-sensitive high-order dependencies that cannot be captured by mean or covariance alone. Such cumulant-driven differences between $c_{2}$ and $c_{3}$ highlight how specific high-order terms introduce distinct learning dynamics. The results validate that networks initially capture low-order moments, but progressively exploit these high-order statistics as training advances. Nonzero cumulants such as skewness and kurtosis provide information unavailable in Gaussian-equivalent data, enabling an improvement in generalization. This progressive utilization of high-order statistics is referred to as ``distributional simplicity bias" in the literature ~\cite{doi:10.1073/pnas.2201854119, 10.5555/3618408.3619607, PhysRevResearch.5.033177, belrose}, which emerges naturally within our framework.

\subsection{Experiments on Fashion-MNIST}
\label{sec:mnist}
To demonstrate that our framework extends beyond synthetic settings to real-world scenarios, we pretrain our data model \eqref{final_data_model} to produce samples from the \textsc{Fashion-MNIST} dataset. Afterwards, to explicitly analyze the role of high-order cumulants, we expand the nonlinearity $\sigma = \tanh$ using Hermite expansion as in \eqref{final_data_model}, truncated at the fifth order ($\ell = 5$). Since $\tanh(\cdot)$ is an odd function, even coefficients vanish ($c_0=c_2=c_4=0$). The remaining coefficients are obtained from the Hermite expansion of $\tanh(\cdot)$, with $c_1$ fixed to preserve the first-order moment, while $c_3$ and $c_5$ are varied to control high-order cumulants. In our experiments with our pretrained data model, we observe that the pretrained model can successfully generate samples like the Fashion-MNIST dataset, confirming the conclusion of Proposition~\ref{prop:prop1}.

For the Gaussian vs. non-Gaussian binary classification experiments (Fashion-MNIST counterpart of the experiments in Fig.~\ref {fig:c2c3}), the effect of high-order Hermite coefficients on learning dynamics is illustrated in Fig.~\ref{fig:fashion}. With $c_{3}=c_{5}=0$, the transformation is nearly linear and the distribution remains close to Gaussian. Introducing a nonzero $c_{3}$ adds skewness, while high-order terms further enrich the cumulant structure. These added moments provide statistical information unavailable in Gaussian-equivalent data, leading to stronger generalization effects. Consistent with synthetic results, the results confirm the controllability of non-Gaussianity through the coefficients $c_i$, and the network's progressive exploitation of cumulants.

\section{CONCLUSION}

This work presents a flexible, analytically tractable framework for studying how high-order cumulants shape learning dynamics. Using Hermite expansions, our nonlinear data model affords precise control over cumulants while remaining highly expressive. . Experiments on both synthetic and real-word data validate these properties and reveal moment-wise learning dynamics. We further find that neural networks learn cumulants sequentially, first capturing low-order moments and then incorporating higher-order features over training. These findings underscore the limitations of Gaussian assumptions and offer a general-purpose tool for investigating non-Gaussian data distributions in learning theory.

\newpage
\bibliographystyle{IEEEbib}
\bibliography{refs}

\end{document}